\documentclass[journal]{IEEEtran}
\usepackage{amsmath}
\usepackage{amsfonts}
\usepackage{algorithmic}
\usepackage{algorithm}
\usepackage{array}
\usepackage[caption=false,font=normalsize,labelfont=sf,textfont=sf]{subfig}
\usepackage{textcomp}
\usepackage{stfloats}
\usepackage{url}
\usepackage{verbatim}
\usepackage{graphicx}
\usepackage{cite}
\usepackage{bbding}
\usepackage{color}
\usepackage{xcolor}
\newcommand{\red}[1]{\textcolor{black}{#1}}
\newcommand{\accept}[1]{\textcolor{black}{#1}}

\begin{document}

\title{A Hybrid Adaptive Controller for Soft Robot Interchangeability}
\author{Zixi Chen, Xuyang Ren, Matteo Bernabei, Vanessa Mainardi, Gastone Ciuti,~\IEEEmembership{Senior member,~IEEE,} and Cesare Stefanini,~\IEEEmembership{Member,~IEEE}
\thanks{We acknowledge the support of the European Union by the Next Generation EU project ECS00000017 ‘Ecosistema dell’Innovazione’ Tuscany Health Ecosystem (THE, PNRR, Spoke 4: Spoke 9: Robotics and Automation for Health.) Z. Chen and X. Ren contributed equally to this work. \emph{*Corresponding authors: Gastone Ciuti}
This work has been submitted to the IEEE for possible publication. Copyright may be transferred without notice, after which this version may no longer be accessible.}
\thanks{The authors are with the Biorobotics Institute and the Department of Excellence in Robotics and AI, Scuola Superiore Sant’Anna, 56127 Pisa, Italy.}
}

\maketitle
\begin{abstract}
Soft robots have been leveraged in considerable areas like surgery, rehabilitation, and bionics due to their softness, flexibility, and safety. However, it is challenging to produce two same soft robots even with the same mold and manufacturing process owing to the complexity of soft materials. Meanwhile, widespread usage of a system requires the ability to \red{replace inner components without highly affecting system performance}, which is interchangeability. 
Due to the necessity of this property, a hybrid adaptive controller is introduced to achieve interchangeability from the perspective of control approaches. This method utilizes an offline-trained recurrent neural network controller to cope with the nonlinear and delayed response from soft robots. Furthermore, an online optimizing kinematics controller is applied to decrease the error caused by the above neural network controller. 
Soft pneumatic robots with different deformation properties but the same mold have been included for validation experiments. \red{In the experiments, the systems with different actuation configurations and the different robots follow the desired trajectory with errors of \red{$\bf{3.3\pm2.9\%}$} and \red{$\bf{4.3\pm4.1\%}$} compared with the working space length, respectively.} Such an adaptive controller also shows good performance on different control frequencies and desired velocities. \red{This controller is also compared with a model-based controller in simulation.}
This controller endows soft robots with the potential for wide application, and future work may include different offline and online controllers. A weight parameter adjusting strategy may also be proposed in the future.
\end{abstract}
\begin{IEEEkeywords}
Soft robot control, LSTM, Robot kinematics, Hybrid controller, Adaptive controller
\end{IEEEkeywords}

\section{Introduction}
\label{sec1}
\IEEEPARstart{S}{oft} robots show infinite degrees of freedom (DOFs), adaptation to environments, and safety to humans. In this case, they have been exploited in various situations, such as medical applications\cite{14CS}, wearable devices\cite{21ZT}, and navigation\cite{16CL}.
Although soft robots have many advantages, the massive application of such robots is partly stuck by manufacturing errors. 
Many soft robots are made with molds and silicone gels. Such a manufacturing process produces soft robots with variances even with the same mold and manufacturing conditions. 
Moreover, expansion and shrinkage during usage will induce unrecoverable deformation.
Soft robots will age due to the materials after a period, which may also change the deformation properties and robot motion.
For symmetrical robots, it is challenging to build a totally symmetrical robot, and the softness in some directions may be different from the other directions. The phenomenon can be found in the unsymmetrical working space in \cite{19TTc,23MN}.
Because of the above reasons, the modeling and control of a soft robot are challenging, and soft robots fail to be utilized as reliable components for massive applications currently.

\begin{figure}[t]
\centering
\includegraphics[width=3.4in]{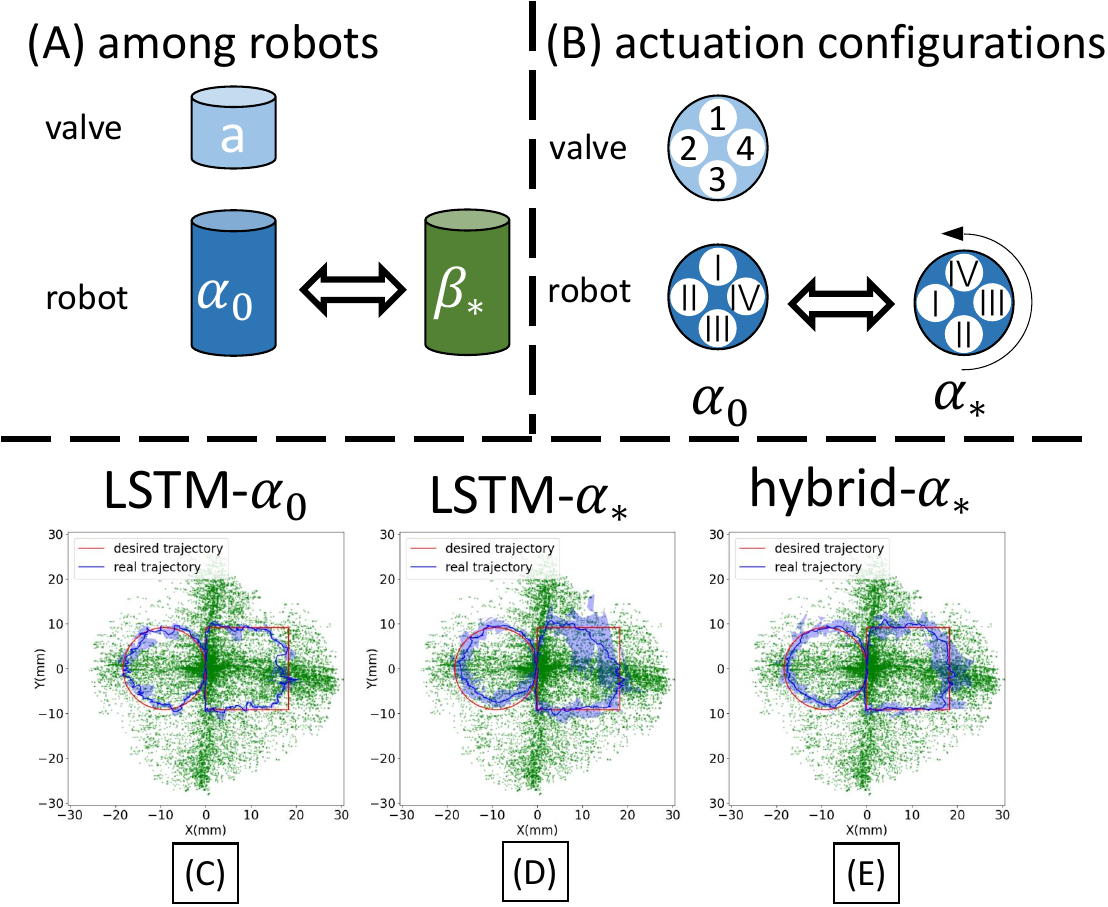}
\caption{Soft robot interchangeability. (A) One soft robot can replace the other robot with the same mold and manufacturing conditions. (B) One symmetrical soft robot can be rotated and connected with valves with different actuation configurations. \red{The LSTM controller is utilized to achieve trajectory following tasks and works well on (C) robot \red{$\alpha_0$} but gets high tracking errors on (D) robot \red{$\alpha_*$}. Meanwhile, the hybrid adaptive controller can achieve the trajectory following task on (E) robot \red{$\alpha_*$.} Light green dots represent the working space. The red and blue lines show the desired and real trajectories, respectively. The light blue area illustrates the error band in \red{nine} trials.}}
\label{fig1}
\end{figure}

Interchangeability, which is the ability of a series of components able to replace each other without changing the system's performance \cite{36JR}, is a significant property of modern manufacturing. For example, bearings of the same size can take the place of the other ones, and they can be used in any direction.
Similarly, as shown in Fig. \ref{fig1}, soft robot interchangeability is required among robots. It is also required among actuation configurations if the robots are symmetrical. Fig. \ref{fig1}-(A) illustrates that in a series of soft robots, one robot can replace the other one without affecting the system performance, and Fig. \ref{fig1}-(B) illustrates that one robot can be connected to valves with any reasonable actuation configurations.
For wide applications, interchangeability is required. However, the manufacturing errors of soft robots mentioned above deteriorate the interchangeability. Hence we aim to enable soft robots to achieve interchangeability with the help of a hybrid adaptive controller.

In this paper, we endow soft robots with interchangeability from the view of control. Recently, neural networks (NNs) have been frequently chosen as the controller of soft robots because they can deal with nonlinear and sequential data, but such data-driven approaches are restricted by the training dataset. Therefore, an online updating controller is required to deal with the mismatch between the training and test robot.

This work leverages an offline trained long short term memory (LSTM) controller and an online optimizing kinematics controller to compose a hybrid adaptive controller and facilitates interchangeability on soft robots. We demonstrate our controller on different actuation configurations and a pair of soft robots which have manufacturing errors. Experimental results show that such robots have different deformation performances but achieve interchangeability thanks to the hybrid adaptive controller. This controller can also fulfill trajectory following tasks under different velocities and control frequencies \red{and outperform a model-based controller in simulated soft arms.}

The contributions of this paper are summarized as follows:
\begin{enumerate}
\item We introduce an offline trained LSTM controller based on data collected from a soft robot with low time and computational cost.
\item We apply an online optimizing kinematics controller to compensate for the errors caused by the difference between the trained and testing robots.
\item Experiments are carried out to prove that soft robots are interchangeable among actuation configurations and robots with the same manufacturing process. The controller is also adaptive to various velocities and control frequencies. \red{This controller performs better than a model-based simulation controller.}
\end{enumerate}

The rest of the paper is structured as follows: Section \ref{sec2} includes works related to neural network controllers and online controllers for soft robots; Section \ref{sec3} introduces our proposed LSTM controller, kinematics controller, and hybrid adaptive controller; Section \ref{sec4} describes the experimental setup, including the robot, experimental devices, and communication diagram; Section \ref{sec5} shows the experimental results which achieve the soft robot interchangeability thanks to the proposed adaptive controller; Section \ref{sec6} summarizes this paper and introduces some possible future works.

\section{Related Work}
\label{sec2}
\subsection{Neural network controllers for soft robots}
\label{sec2.1}
NN is attractive to soft robot scholars because it is composed of nonlinear activation functions, which may be effective approaches to deal with the nonlinear behaviors of soft robots. In 2007, NN is first involved in soft robot research in \cite{07DB}, and furthermore, NN is leveraged for modeling \cite{22XHb} and control\cite{15MG}.

Recently, recurrent neural networks (RNN) have been shown as viable choices for soft robot research. The recurrent structures can cope with time-related data relationships. Therefore, RNN is an effective tool for soft robot control. 
One RNN named nonlinear autoregressive network with exogenous inputs (NARX) is applied in \cite{18TT} for soft robot open loop control to achieve trajectory following tasks. 
In RNN, LSTM shows good performance due to its long term memory. The authors of 
 \cite{22DW} illustrate that LSTM outperforms an analytical model in control tasks. Also, LSTM is employed in \cite{21ZD} on a soft pneumatic finger for contact position and force estimation. LSTM in \cite{23AA} estimates finger motion and contact force, and this paper shows that the estimation accuracy will improve with more sensing channels. The authors of \cite{23LW} demonstrate that LSTM and one other RNN named gate recurrent units can estimate symmetrical soft robot motion under random contact and partial sensor failure.

Although the training datasets provide information for the neural networks, they pose limitations in their application areas meanwhile. In \cite{23MN}, NN trained with simulation dataset shows unsatisfying control performance on the real robot. To perform adaptation between different actuation configurations and different robots, an online adjusting controller is involved in this work.

\subsection{Online controllers for soft robots}
\label{sec2.2}
Statistical controllers, like the Gaussian mixture model \cite{19BY} and Gaussian process regression\cite{22ZT}, have low data amount requirements compared with NN controllers. In this case, they can update online and hence are robust to model misalignment and unpredicted noises.
Kalman filter is also a popular choice for online-updating controllers. A Kalman filter is applied in \cite{18ML} for Jacobian estimation and builds an optimal controller based on the estimated Jacobian matrix.

Jacobian methods have also been employed in soft robot control as online updating controllers. Such approaches suppose the differences between end positions and actions are linear. 
The Jacobian method is first introduced in \cite{14MY}. The actuation variables are decided by an optimization problem considering the Jacobian matrix as a constraint. Instead, the inverse Jacobian matrix for control is applied in \cite{20XW}. Such a simple approach can also be leveraged to control contact force with a force-displacement model \cite{16MY}.

Due to the linear assumption and noisy response, the Jacobian method requires a high control frequency, but the kinematics controller applied in this paper works well even at a low frequency.
Compared with online learning NN which requires a data collection process and cannot update every time step \cite{23MN}, the kinematics controller can update with the same frequency as the control strategy. The summary of more RNN and online controllers applied to soft robot control can be found in \cite{23CL,23ZCb}.

\section{Method}
\label{sec3}
\begin{figure}[!ht]
\centering
\includegraphics[width=3.4in]{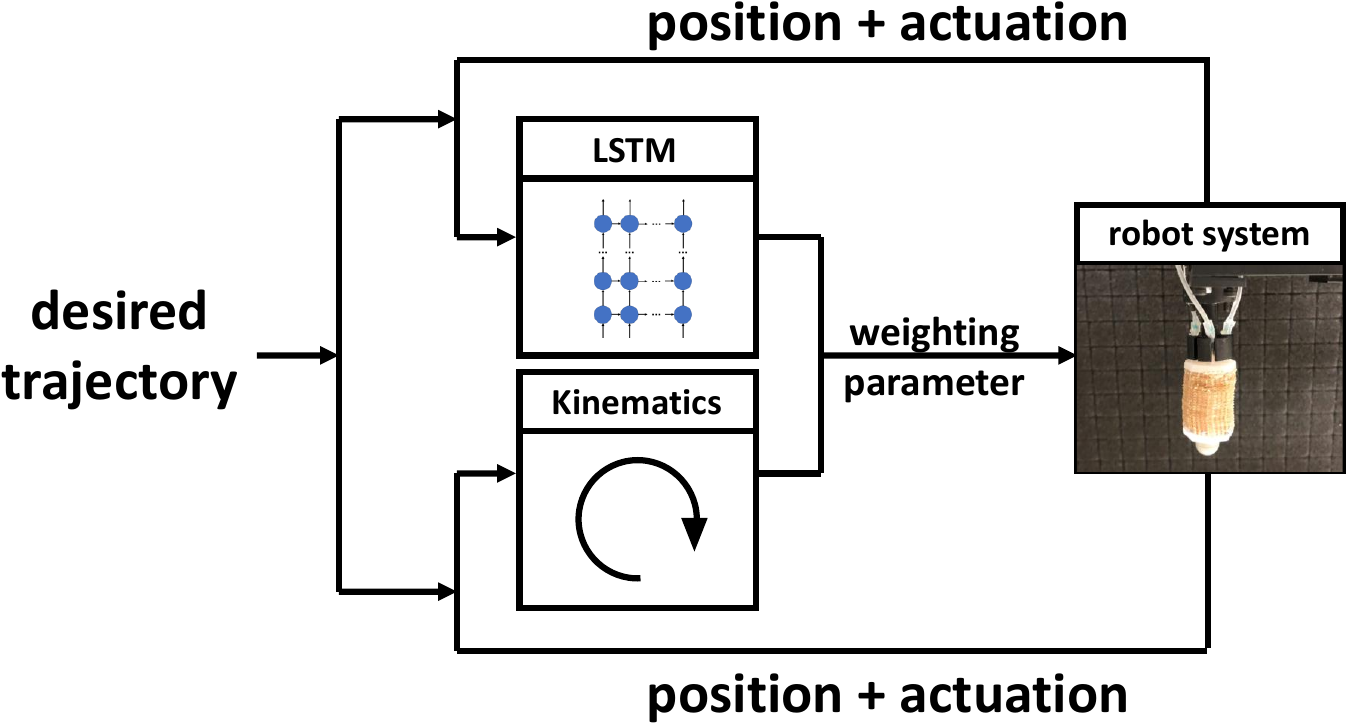}
\caption{Control strategy diagram. The sensor in the robot system provides the end position for the controller. The desired trajectory and actuation in the last steps are also sent to the controller. The kinematics controller updates based on the dataset collected currently, and both the kinematics controller and LSTM estimate actuation for the next step. A weight parameter is used to weigh the estimated actuation variables from those two controllers, and the weighted actuation is sent to the real robot system to achieve the trajectory following tasks.}
\label{fig2}
\end{figure}

This section introduces the hybrid adaptive controller exploited in this paper. We train an LSTM network offline at first and propose an online optimizing kinematics controller for adaptation. The estimated results from these two controllers are weighed and executed finally. The control strategy diagram is depicted in Fig. \ref{fig2}.

\subsection{LSTM Controller}
\label{sec3.1}
In an RNN network, the information from the previous steps can be taken into consideration thanks to the hidden states $h$, and LSTM can select proper information to forget and remember with the cell states $C$. For the cell representing step $t$, the calculation is

\begin{equation}
\label{eq2_1}
\begin{split}
f_t=&sig(W_f \cdot [h_{t-1},x_t]+b_f),\\
i_t=&sig(W_i \cdot [h_{t-1},x_t]+b_i),\\
C_t=&f_t\times C_{t-1}+i_t\times tanh(W_c \cdot [h_{t-1},x_t]+b_c),\\
o_t=&sig(W_o \cdot [h_{t-1},x_t]+b_o),\\
h_t=&o_t\times tanh(C_t),
\end{split}
\end{equation}
where $f_t, i_t, C_t, o_t,$ and $h_t$ denote the forget gate, input gate, cell state, output gate, and hidden state of step $t$, respectively. $sig$ is the sigmoid function. $x_t$ is the input of step $t$. $W_*$ and $b_*$ represent the weight and bias parameters of the corresponding calculation, and $\times$ is the Hadamard product operator in these equations.

To utilize an LSTM network as a controller, a network is trained with a collected dataset offline. The previous actuation values and positions are employed as inputs, and the current actuation values are employed as the output. The training process can be summarized as

\begin{equation}
\label{eq2_2}
\begin{split}
a_{LSTM,t} = LSTM(p_{t+1};&p_{t},p_{t-1},\dots;\\&a_{t-1},a_{t-2},\dots),
\end{split}
\end{equation}
where $p_t\in R^{2\times1}$ and $a_t\in R^{2\times1}$ denote the end position and actuation value at step $t$, and $a_{LSTM,t}\in R^{2\times1}$ represents the actuation value estimated by LSTM controller at step $t$. In this paper, we employ the 2D position of the robot end to represent robot states, and there are two independent actuation variables.
During the control process, the end position from the desired trajectory will take the place of $p_{t+1}$, and LSTM will estimate the corresponding actuation values according to the desired trajectory and robot motion history. The hyperparameter choices like the previous step number, layer number, and hidden state feature number will be introduced in Subsection \ref{sec5.1}.

\subsection{Kinematics Controller}
\label{sec3.2}
The motivation behind utilizing an online updating controller lies with the observation that neural network controllers are constrained by the training dataset and fail to adapt to new robots. Therefore, the kinematics controller is leveraged to compensate for the mismatch between the test robot and the robot training the LSTM. Due to its low data amount requirement, the kinematics controller can learn the robot dynamics online every time step. The kinematics matrix updates by solving the following optimization problem:

\begin{equation}
\label{eq2_3}
\begin{split}
\red{\min_{K}}\ &\red{{\Vert \triangle \hat{p} \Vert}_2}\\
\red{s.t.}\ &\red{\triangle \hat{p} = P-KA}\\
\end{split}
\end{equation}
where $K\in R^{2\times2}$ is the optimized kinematics matrix, $P=[p_{t}\ p_{t-1}\ ...\ p_{t-4}]\in R^{2\times5}$, $A=[a_{t-1}\ a_{t-2}\ ...\ a_{t-5}]\in R^{2\times5}$. $\triangle \hat{p}$ denotes the estimation loss of the kinematics matrix $K$. \red{We apply the identity matrix $I_{2\times2}$ as the initial kinematics matrix in the beginning, and the optimized matrix of this time step will be utilized as the initial guess of the next time step.}

To utilize the kinematics method as a controller, similar to \cite{20YW}, we employ the other optimization problem to calculate the actuation variables. The optimization problem is

\begin{equation}
\label{eq2_4}
\begin{split}
\min_{a_{K,t}}\ &{\Vert \triangle {p} \Vert}_2\\
s.t.\ &\triangle {p} = p_{tar,t+1}-K{a_{K,t}}\\
\end{split}
\end{equation}
where $p_{tar,t+1}\in R^{2\times1}$ is the target position of the step $t+1$ and $a_{K,t}\in R^{2\times1}$ denotes the actuation variable estimated by the kinematics controller. We may include the inverse matrix for controlling directly or the other online controller in the future.

\subsection{Hybrid Controller}
\label{sec3.3}
The offline trained LSTM controller and online optimizing kinematics controller are utilized to cope with the nonlinear and delayed response from soft robots while adapting the real test robots. A weighting parameter $w$ is applied to compose this hybrid controller, which can be summarized as

\begin{equation}
\label{eq2_5}
\begin{split}
a_{hy,t} = w\cdot a_{K,t}+ (1-w)\cdot a_{LSTM,t},
\end{split}
\end{equation}
where $a_{hy,t}$ represents the actuation value estimated by the hybrid controller at step $t$. In this work, $w$ is set as $0.1$ after trials and errors and may be adjusted online inspired by the Kalman filter \cite{18DL} in future work.

\section{Experimental Setup}
\label{sec4}

\begin{figure}[!ht]
\centering
\includegraphics[width=3.4in]{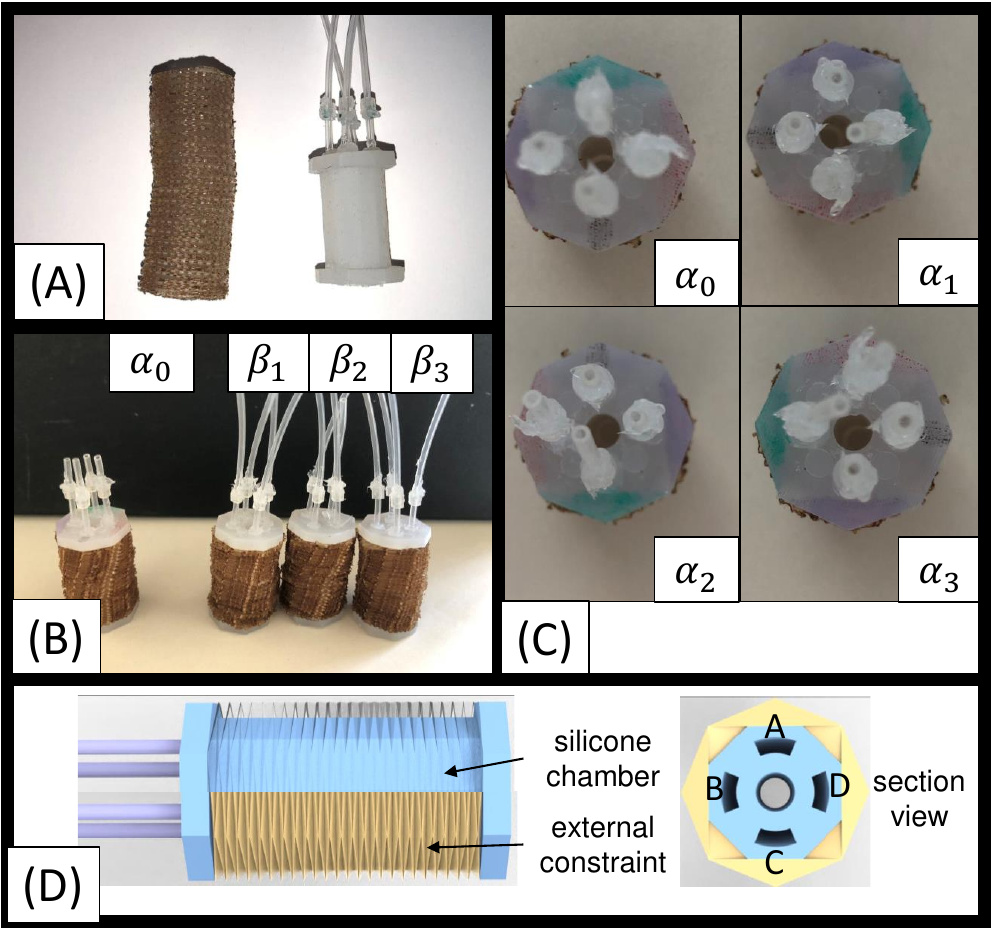}
\caption{(A) A robot is composed of an origami structure and a silicone structure containing four chambers. \red{(B) Four different soft robots from the same mold, which are $\alpha_0$, $\beta_1$, $\beta_2$, and $\beta_3$. (C) Four different actuation configurations of the same robot, which are $\alpha_0$, $\alpha_1$, $\alpha_2$, and $\alpha_3$. (D) Section view of a soft robot. There are four chambers inside a robot, and an external constraint limits the radial deformation.}}
\label{fig3}
\end{figure}

In this work, \red{four} soft pneumatic robots \red{made with the same mold and manufacturing conditions and four actuation configurations of one robot} are included, as shown in Fig. \ref{fig3}\red{-(B) and (C)}. \red{$\beta_*$ represents $\beta_1$, $\beta_2$, and $\beta_3$, and $\alpha_*$ represents $\alpha_1$, $\alpha_2$, and $\alpha_3$.} \red{Four chambers are placed at an interval angle of $90^\circ$ along the circumferential direction in each robot. The robot could realize the two-dimensional motion easily with this kind of symmetric layout. The main parts of these robots are made of \red{Ecoflex 00-50} (Smooth-On, Macungie, PA) and the ends are made of Dragon Skin 30 (Smooth-On, Macungie, PA.) One origami structure packs each robot to constrain radial deformation and promote elongation. The origami structures of \red{these robots} are made of Polyvinyl Chloride (PVC)\red{, as shown in Fig. \ref{fig3}-(A)}. For simplicity, two symmetrical chambers are set into a pair, and only one chamber in each pair will be actuated in a time step according to the actuation values. In this case, two actuation values are applied to control a soft robot with two chamber pairs, and end positions on a 2D plane are applied to build a robot motion dataset. For example, if the pressure of the left and front chambers are controlled to improve, the end will move to the right and back. Due to the silicone and origami structure material constraints, the maximal actuation pressure is about 0.5 bar.}

\begin{figure}[!ht]
\centering
\includegraphics[width=3.4in]{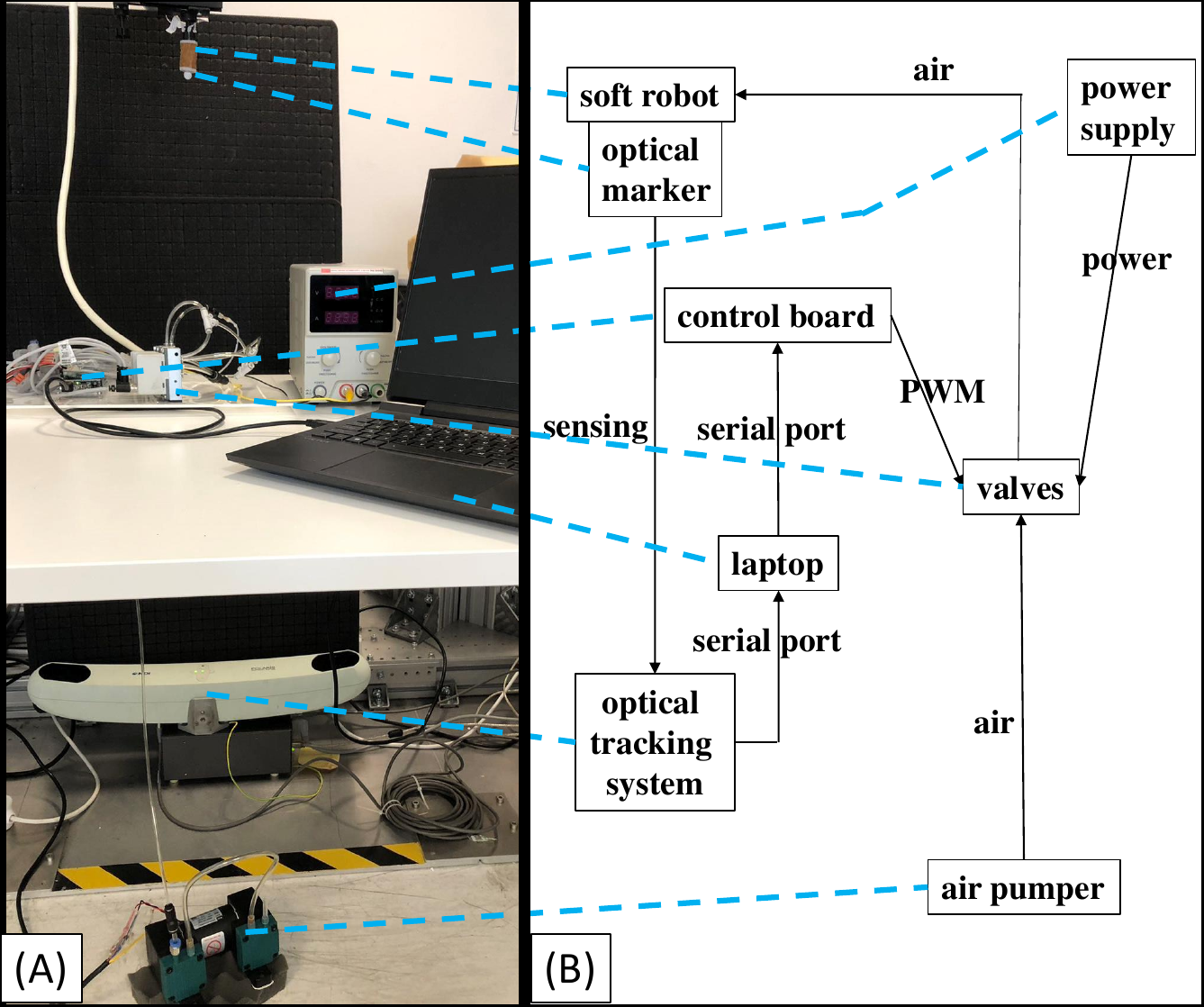}
\caption{\red{(A) Experiment devices and (B) hardware communication diagram.} A marker is fixed on the end of the soft robot. The robot is actuated by four valves which are controlled by a control broad. A personal computer is utilized for communicating with the sensing system and control board.}
\label{fig4}
\end{figure}

\begin{figure}[!ht]
\centering
\includegraphics[width=3.4in]{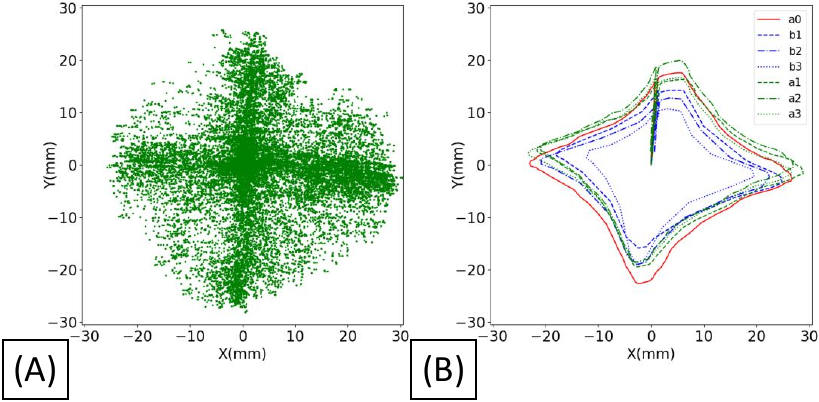}
\caption{\red{(A) Working space. The length of the working space side is $60.94mm$. (B)Trajectories from $\alpha_0$, $\alpha_*$, and $\beta_*$. They are actuated with the same actuation sequence but produce different trajectories.}}
\label{fig5}
\end{figure}

The experimental devices \red{and communication diagram} are shown in \red{Fig. \ref{fig4}-(A) and (B).} 
NDI Polaris Spectra, an optical measurement system, is applied for end position sensing. Its error is $0.25mm$ RMS, which is lower than that of NDI Aurora ($0.48mm$ RMS), the EM tracking system widely applied in soft robot research. 
One robot is fixed vertically on a base, and a reflection marker is fixed on the end of the robot. The marker is inside the sensing area of NDI Polaris in all the experiments.
Four chambers of the robot are connected to four valves (Camozzi K8P-0-E522-0) which are under the control of the Arduino MEGA control board. 
A laptop \red{(Ubuntu 20.04, CPU i5-12500H, and RTX 3050)} is leveraged for network training and control strategy implementation.
It communicates with both the sensing system and control board using serial communication in ROS.

\red{Fig. \ref{fig5}-(A) shows $20000$ samples collected from robot $\alpha_0$, which are shown as green dots and can be seen as the working space of the robot.} This dataset is collected with a frequency of about $3.3$Hz, and such frequency is applied to the control and sensing system in the following experiments. During the training process, the robot is driven by random actuation variables. The actuation value differences between two adjacent steps are constrained to generate a smooth motion. It takes roughly 100 minutes to collect this dataset. The original length of extendable chambers is about $25mm$, and the working space area of a robot is \red{$60.94mm\times60.94mm$}. It is evident that although the robot is expected to be totally symmetrical, the working space is asymmetrical. The left and \red{top} parts of the working space are more challenging to reach than the others\red{, which shows that there are manufacturing errors among actuation configurations. Fig.\ref{fig5}-(B) shows trajectories from different robots based on the same actuation sequence. The average error among $\alpha_0$ and $\alpha_*$ is $1.52mm$, and the average error among $\alpha_0$ and $\beta_*$ is $5.21mm$
Although they are from the same molds, they generate different trajectories, showing manufacturing errors among robots.}

\section{Experimental Results}
\label{sec5}
\subsection{LSTM Controller Performance}
\label{sec5.1}

Based on the dataset, we train an LSTM network as the offline controller with PyTroch \cite{19AP}. The dataset shown in Fig. \ref{fig5}-(A) is divided into a training dataset ($70\%$), a validation dataset ($10\%$), and a test dataset ($20\%$.) The validation dataset is exploited by the early stopping strategy during training to prevent overfitting, and we test the trained LSTM with the test dataset. According to Subsection \ref{sec3.1}, the actuation variables and positions from several previous steps are employed as input, and the output is the estimated actuation variable for the current time step.

To obtain a proper hyperparameter combination, various combinations are tried. \accept{We train NNs with several hyperparameter combinations on the training dataset and test them on the test dataset. The test errors are shown in Table \ref{table1}.} The network with the lowest error is considered the best controller before real experiments. In this case, $10$ previous time steps, $4$ layers, hidden state size $128$, and dropout rate $0.1$ are adopted for the following experiments. Except for this combination, the other choices also provide low errors; hence careful and time-consuming fine-tuning is not necessary for massive applications. 
It takes roughly $100$ minutes to collect the $20000$ data points and under \red{$5$} minutes to train an LSTM network. The collection and training processes are far shorter than those in \cite{18IK} (about 10 hours), which applies time-consuming reinforcement learning.

\begin{table}[!ht]
\caption{LSTM Test Errors}
\centering
\begin{tabular}{l|l }
layer number-previous time step-\\hidden state size-dropout rate & Test Error \\
\hline
{4-10-128-0.1} & \red{$\bf{1.84\%}$}\\
{2-10-128-0.1} & \red{$2.16\%$}\\
{4- 6-128-0.1} & \red{$2.75\%$}\\
{4-10- 64-0.1} & \red{$2.98\%$}\\
{4-10-128-0.3} & \red{$3.20\%$}\\
\end{tabular}
\label{table1}
\end{table}

\begin{figure}[!ht]
\centering
\includegraphics[width=3.4in]{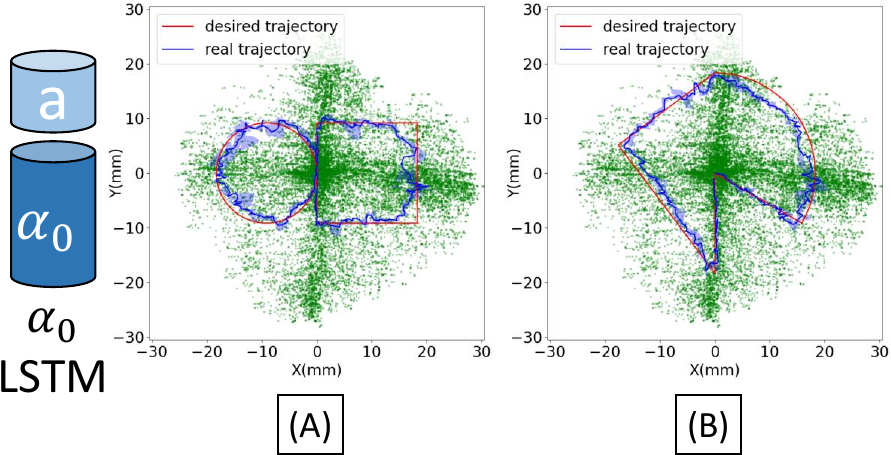}
\caption{Trajectory following results on robot \red{$\alpha_0$}. LSTM tracking performance on \red{(A) A and (B) B} trajectories. Light \red{green} dots represent the working space. The red and blue lines show the desired and real trajectories, respectively. The light blue area illustrates \red{the error band} in \red{three trials.}}
\label{fig6}
\end{figure}

\red{Two trajectories are proposed as the following targets for different controllers as shown in Fig. \ref{fig6}-(A) and (B).
The first trajectory is composed of one circle and one square. The second trajectory is composed of one triangle and one curve. These two trajectories contain acute angles, right angles, obtuse angles, straight lines, and curved lines.}
Each trajectory is divided into 400 positions, which are employed as the target positions in the following tasks.
These two trajectories are named \red{as \textit{A} and \textit{B}} in this paper.

The robot \red{$\alpha_0$}, which generates the dataset, is controlled to follow these trajectories with the LSTM controller, and the performance is shown in Fig. \ref{fig6}-(A) and (B). The distance error to working space length (\red{$60.94mm$}) ratio is applied to represent the trajectory following errors in this paper. The following errors and \red{standard deviations} are shown in Table \ref{table2}, which illustrates that the LSTM controller fulfills trajectory following performance perfectly with the robot \red{$\alpha_0$}.

\begin{table}[!ht]
\caption{Trajectory Following Errors}
\centering
\begin{tabular}{l|l l}
Robot-Controller & \red{A} & \red{B}\\
\hline
\red{$\alpha_0$-LSTM}   & \red{$3.4\pm2.7\%$} & \red{$3.3\pm1.8\%$}\\
\hline
\red{$\alpha_*$-LSTM}   & \red{$4.3\pm4.0\%$} & \red{$4.5\pm3.2\%$}\\
\red{$\alpha_*$-Hybrid} & \red{$3.3\pm2.9\%$} & \red{$3.4\pm2.5\%$}\\
\hline
\red{ $\beta_*$-LSTM}   & \red{$6.8\pm5.9\%$} & \red{$8.1\pm6.1\%$}\\
\red{ $\beta_*$-Hybrid} & \red{$4.3\pm4.1\%$} & \red{$4.9\pm3.6\%$}\\
\end{tabular}
\label{table2}
\end{table}

\begin{figure}[!ht]
\centering
\includegraphics[width=3.4in]{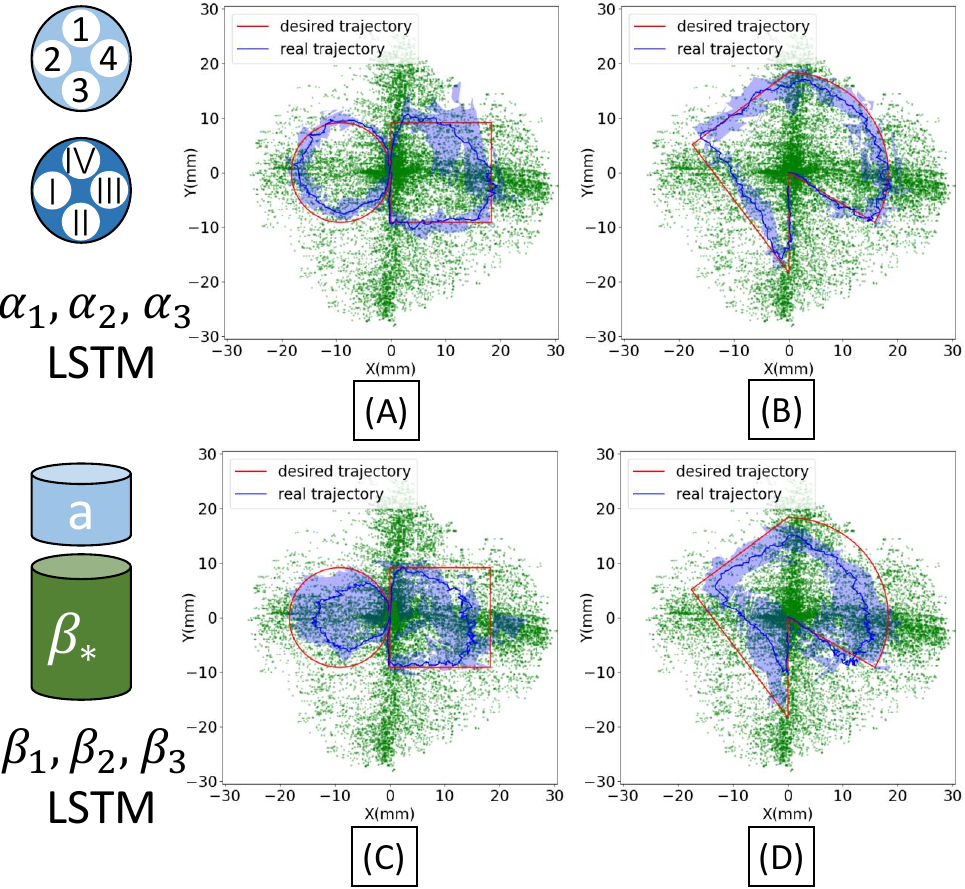}
\caption{LSTM controller performance on (A) robot \red{$\alpha_*$, A trajectory}; (B) robot \red{$\alpha_*$, B trajectory}; (C) robot \red{$\beta_*$, A trajectory}; (D) robot \red{$\beta_*$, B trajectory}. Light \red{green} dots represent the working space. The red and blue lines show the desired and \accept{the average of the }real trajectories, respectively. The light blue area illustrates \red{the error band} in \red{nine trials.}}
\label{fig7}
\end{figure}

\red{Then the robot $\alpha_0$ with different actuation configurations, represented by $\alpha_*$ in Fig. \ref{fig3}-(C), and the robots $\beta_*$ in Fig. \ref{fig3}-(B) are controlled to follow the trajectories.}
Fig. \ref{fig7} demonstrates that these hardware implementations obtain higher errors than the robot \red{$\alpha_0$} under the control of LSTM. Their errors and \red{standard deviations} can be found in Table \ref{table2}. \accept{Each subfigure in Fig. \ref{fig7} includes nine experiments, considering three configurations or robots and three trials on each configuration or robot.}
LSTM obtains both high errors and \red{standard deviations} on \red{actuation configurations $\alpha_*$ and robots $\beta_*$}, which illustrates that this data-driven approach is not effective and robust under the robot beyond the training dataset.

\begin{figure}[!ht]
\centering
\includegraphics[width=3.4in]{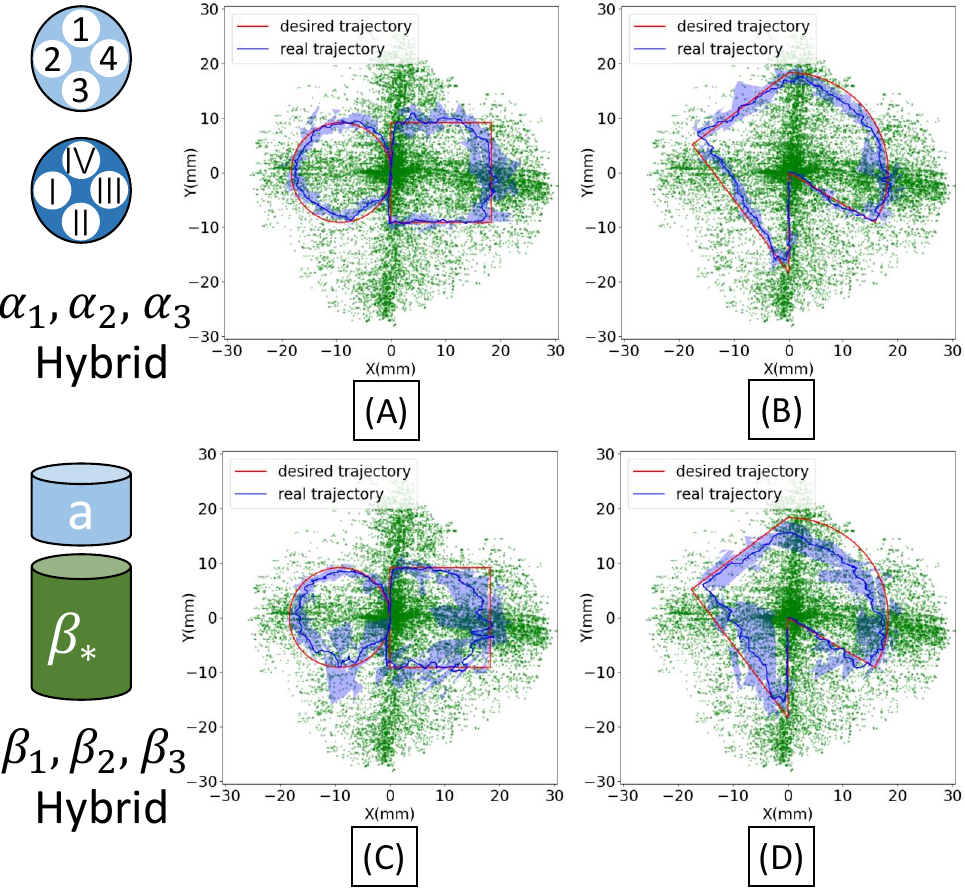}
\caption{Hybrid controller performance on (A) robot \red{$\alpha_*$, A trajectory}; (B) robot \red{$\alpha_*$, B trajectory}; (C) robot \red{$\beta_*$, A trajectory}; (D) robot \red{$\beta_*$, B trajectory}. Light \red{green} dots represent the working space. The red and blue lines show the desired and real trajectories, respectively. The light blue area illustrates \red{the error band} in \red{three trials.}}
\label{fig8}
\end{figure}

\subsection{Hybrid Controller Performance}
\label{sec5.2}
To compensate for the difference between the test robot \red{$\alpha_*$, $\beta_*$ and robot $\alpha_0$}, the online optimizing kinematics controller is involved for control according to Eq. \ref{eq2_5} with the help of PyTroch \cite{19AP}. Fig. \ref{fig8} illustrates that the hybrid controller decreases following errors and achieves interchangeability. The errors and \red{standard deviations} can be found in Table \ref{table2}. 
\red{Considering the high LSTM trajectory following errors of $\beta_3$ (over $10\%$), we improve the weight parameter $w$ from $0.1$ to $0.5$, and the errors of the hybrid controller decrease to under $6\%$.}
\accept{The weight change is only applied in $\beta_3$, and $0.1$ is still applied in the other robots. Fig. \ref{fig8}-(C), (D) include results of robot $\beta_3$ with weight $0.5$ and the other robots with weight $0.1$.}
\red{This controller achieves the smallest errors on configuration $\alpha_3$, trajectory B, which is $2.8\%$.
The adaptive controller decreases the errors of trajectory following tasks.}

\begin{figure}[!ht]
\centering
\includegraphics[width=3.4in]{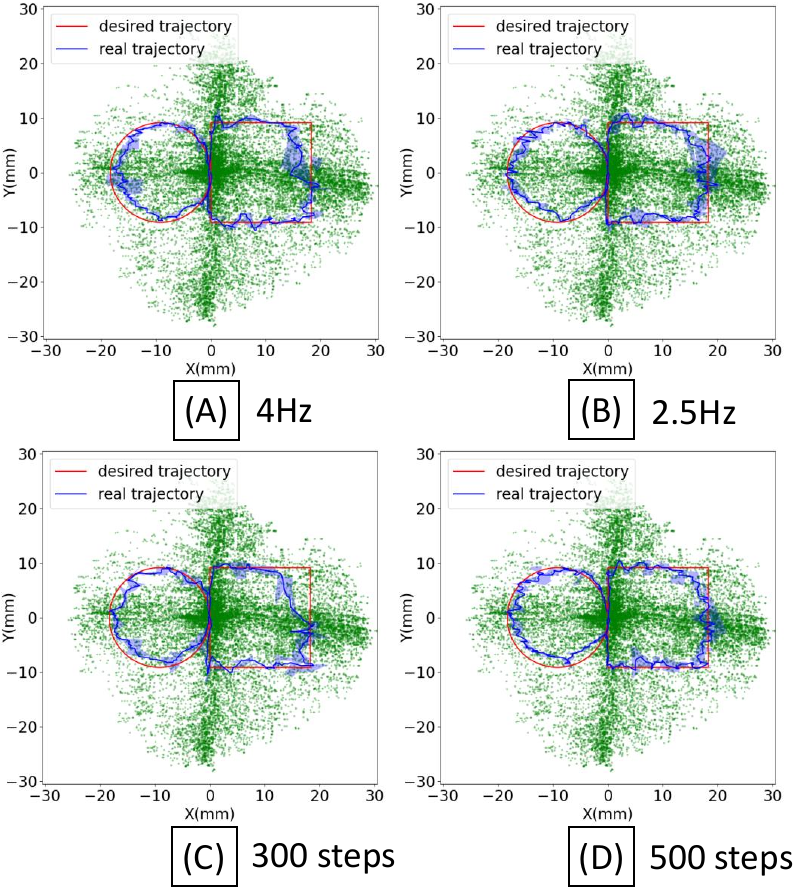}
\caption{\red{Hybrid controller performance on robot $\alpha_*$ with (A) $4$Hz, (B) $2.5$Hz, (C) $300$ time steps, (D) $500$ time steps and trajectory A.} Light \red{green} dots represent the working space. The red and blue lines show the desired and real trajectories, respectively. The light blue area illustrates \red{the error band} in \red{three trials.}}
\label{fig9}
\end{figure}

To demonstrate the adaptation of the hybrid controller, we exploit the controller in experiments with various control frequencies and velocities on \red{the robots $\alpha_*$ and $\beta_*$} and the trajectory \red{A and B}. \red{To test the adaptation of control frequency, we change the frequency from $3.3$Hz to \red{$2.5$Hz} and $4$Hz. In this case, the time step is about $0.4$s and $0.25$s.
To test the adaptation of velocity, the time step number changes from \red{$400$ to $300$ and $500$}. \accept{In the experiments of Fig. \ref{fig7}, the whole desired trajectory is discretized into $400$ positions, and here we discretize it into $300$ and $500$ positions.}
In this case, the desired velocity changes indirectly.}
\accept{The errors and derivations can be found in Table \ref{table3}, and the following performance of $\alpha_*$ and trajectory A is shown in Fig. \ref{fig9}.}
Overall, the adaptive controller can fulfill these tasks with relatively low errors. The LSTM network is trained with data from $3.3$Hz, but the adaptive hybrid controller obtains good performance even on different frequencies.

\begin{table}[!ht]
\caption{Following Errors for Adaptation Tasks}
\centering
\begin{tabular}{l|l l}
Robot-Experimental situation &\red{A} &\red{B} \\
\hline
\red{$\alpha_*$-4Hz}       & \red{$3.0\pm2.1\%$} & \red{$3.5\pm2.1\%$}\\
\red{$\alpha_*$-2.5Hz}     & \red{$3.3\pm2.8\%$} & \red{$3.8\pm2.9\%$}\\
\red{$\alpha_*$-300 steps} & \red{$4.2\pm3.7\%$} & \red{$4.1\pm2.2\%$}\\
\red{$\alpha_*$-500 steps} & \red{$2.6\pm2.3\%$} & \red{$3.7\pm2.9\%$}\\
\hline
\red{$\beta_*$-4Hz}        & \red{$4.4\pm4.0\%$} & \red{$5.0\pm3.2\%$}\\
\red{$\beta_*$-2.5Hz}      & \red{$4.5\pm4.1\%$} & \red{$5.1\pm4.0\%$}\\
\red{$\beta_*$-300 steps}  & \red{$4.9\pm3.9\%$} & \red{$6.2\pm4.8\%$}\\
\red{$\beta_*$-500 steps}  & \red{$3.8\pm3.7\%$} & \red{$6.2\pm5.2\%$}\\
\end{tabular}
\label{table3}
\end{table}

\subsection{Ablation experiments}
\label{sec5.3}
An ablation study is applied to explore the function of a certain component in a system by running the system without that part. Such an experiment is widely used in machine learning\cite{19RM}. In this paper, ablation experiments are carried out for the hybrid controller to demonstrate the necessity of each part of the controller. The performance of the LSTM controller has been shown in Fig. \ref{fig7}. This component gains high performance in the original robot but may fail in new robots. To some extent, it can be seen as a global controller because it does not rely highly on feedback and has the potential to be used everywhere in the working space.

\begin{figure*}[!ht]
\centering
\includegraphics[width=7.1in]{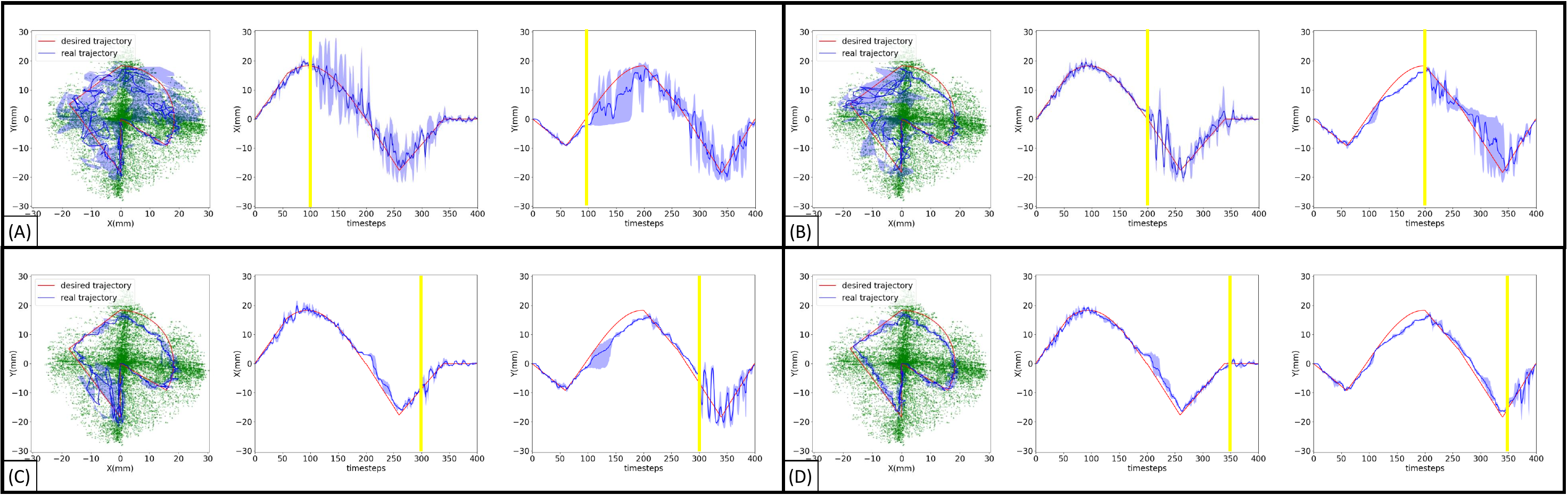}
\caption{\red{Kinematics controller performance on the robot $\alpha_*$. The trajectories, end positions on the x-axis and y-axis are shown.} The hybrid controller is employed at first, and the kinematics controller starts at (A) 50th, (B) 100th, (C) 200th, and (D) 300th, respectively. Light \red{green} dots represent the working space. The yellow line shows the start time step of the kinematics controller. The red and blue lines show the desired and real trajectories, respectively. The light blue area illustrates \red{the error band} in \red{three trials.}}
\label{fig10}
\end{figure*}

Considering the kinematics controller, Fig. \ref{fig10} shows its performance on the robot \red{$\alpha_1$} and the trajectory \red{$B$}. The hybrid controller is utilized at first, and the kinematics replaces it \red{at $100th, 200th, 300th,$ and $350th$} time steps, which is achieved by changing the weight parameter $w$ from $0.1$ to $1$ in Eq. \ref{eq2_5}. The following errors and derivations are shown in Table \ref{table4}.

\begin{table}[!ht]
\caption{Kinematics Controller Following Errors with $\alpha_1$ and Trajectory B}
\centering
\begin{tabular}{l|l}
Switch step & Error \\
\hline
\red{100}  & \red{$8.7\pm7.5\%$}\\
\red{200}  & \red{$6.1\pm5.6\%$}\\
\red{300}  & \red{$5.1\pm3.9\%$}\\
\red{350}  & \red{$3.9\pm2.5\%$}\\

\end{tabular}
\label{table4}
\end{table}

The kinematics controller shows high following errors in the first three tasks. After switching to the kinematics controller, the robot still tries to follow the trajectories in several time steps but starts to oscillate after a while, which illustrates that it is just a local controller and may fail considering a global task with turning. However, the fourth task, as shown in Fig. \ref{fig10}-(D), is achieved perfectly, and the error (\red{$3.9\%$}) is even lower than that of the hybrid controller on $\alpha_1$ (\red{$4.1\%$}). This error reduction demonstrates that the kinematics controller outperforms the hybrid controller, and the weighting parameter $1$ is a better choice for some simple straight-line following tasks. Meanwhile, the weighting parameter $w$ changes from $0.1$ to \red{$0.5$} for complex tasks like the robot \red{$\beta_3$} trajectory following. In this case, an online weight-adjusting strategy may improve the overall controller performance.

\subsection{\red{Simulation experiments}}
\label{sec5.4}
\begin{figure}[!ht]
\centering
\includegraphics[width=3.4in]{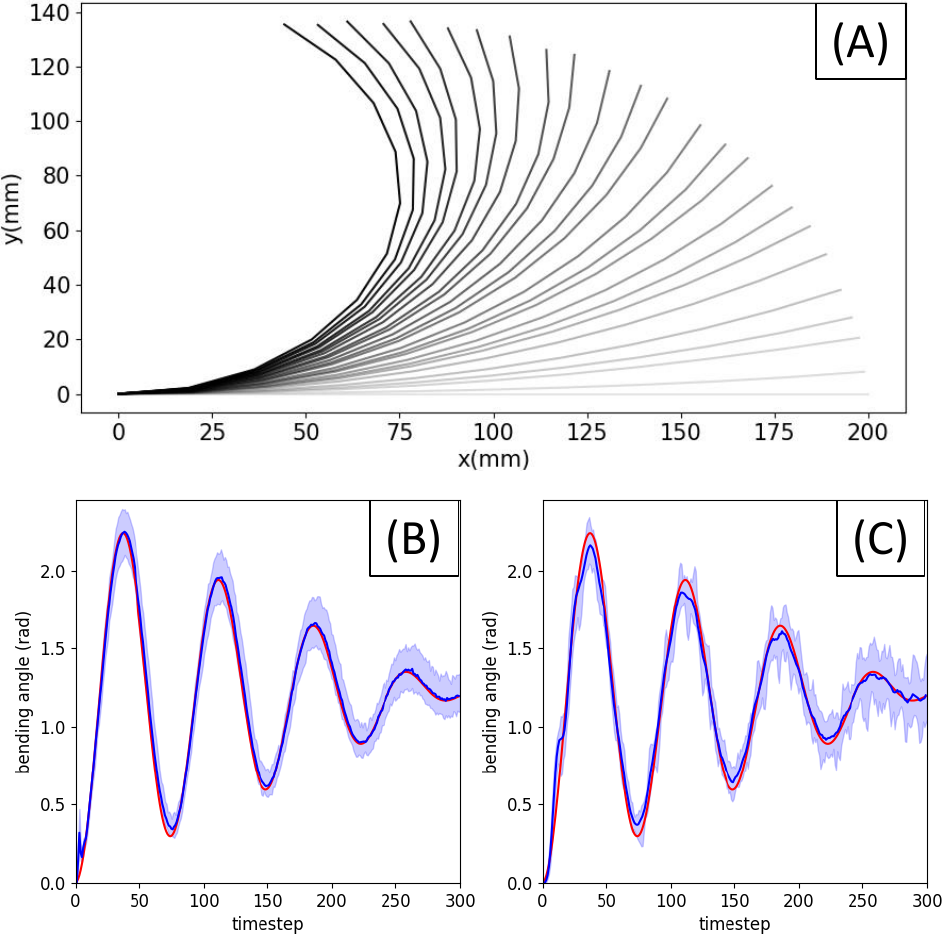}
\caption{\red{(A) Diagram of soft arms from 0 to 2.4 rads. Tracking performance with (B) hybrid adaptive controller and (C) CC controller. The red and blue lines show the desired and \accept{the average of the} real trajectories, respectively. The light blue area illustrates the error band in 25 trials.}}
\label{fig11}
\end{figure}

\red{Besides real pneumatic four-chamber robots, we also apply our hybrid adaptive controller to simulated 1-DOF soft arms and compare it with a model-based controller. The soft arm simulation is built based on PyElastica\cite{18MG}, as shown in Fig. \ref{fig11}-(A). The working space in these experiments is 0-$2.4$ rads and data in $1000$ time steps is collected for LSTM training.
Based on the Constant Curvature (CC) model\cite{18CSb} and  pseudo-rigid-body model\cite{19ZT}, we build the soft robot dynamics model as
\begin{equation}
\label{eq5_1}
\begin{split}
B\Ddot{q}+C\dot{q}+Kq=\tau,
\end{split}
\end{equation}
where $q$ is the bending angle, $B, C,$ and $K$ are the robot’s inertia, damping, and stiffness factor, and $\tau$ is the control input. According to \cite{18CSb}, the controller is 
\begin{equation}
\label{eq5_2}
\begin{split}
\tau = B\Ddot{\Bar{q}}+C\dot{\Bar{q}}+K\Bar{q}+I\int{(\Bar{q}-q)},
\end{split}
\end{equation}
where $\Bar{q}$ is the desired angle trajectory, and $I$ is the gain of the integral action.}

\red{To test the adaption of controllers, the parameters of these two controllers are optimized on a soft arm with Young's modulus $10$ kPa and Possion ratio $0.5$. These deformation parameters have been applied in \cite{23HC}. Then, we test them on total $25$ robots with Young's modulus $5, 7.5, 10, 12.5, 15$ kPa and Possion ratio $0.25, 0.375, 0.5, 0.625, 0.75$, as shown in Fig. \ref{fig11}-(B) and (C). The test errors and standard deviations of the hybrid adaptive controller and CC controller are $2.8\pm3.4\%$ and $3.0\pm3.9\%$. Therefore, our controller outperforms a model-based controller and shows adaptive control ability on various robots.
}

\section{Conclusion and Discussion}
\label{sec6}
This work aims to achieve soft robot interchangeability with the help of an adaptive controller. An offline trained LSTM controller and an online optimizing kinematics controller compose this hybrid adaptive controller, and such a controller can obtain satisfying control performance on different actuation configurations and robots. These results demonstrate that such a controller is an effective and time-saving solution for soft robot control with different robots, trajectories, and frequencies. Furthermore, it paves the way for the massive application of soft robots in industry.

In future work, the other sequence-related neural network, like gated recurrent units, may take the place of LSTM. Also, some other online learning approaches, like the transfer function controller, the Jacobian method, and Gaussian mixture regression, may replace the kinematics controller. \red{We may include different numbers of timesteps in the kinematics controller to explore the influence of this parameter.}
\red{Another fundamental issue that must not be neglected is the choice of the weight parameter. An online updating strategy may be proposed for adjusting the kinematics weight inspired by the Kalman gain updating strategy in the Kalman filter. }
We also plan to achieve the interpretability of neural networks to some extent from the view of mechanics and robotics like \cite{23JL}.
\red{Owing to its ability to adjust to various robots, this control strategy may also be used on robots with different physical properties and complex robot systems like the reconfigurable modular robot system\cite{10HK}.}
\bibliographystyle{IEEEtran}
\bibliography{IEEEabrv,references}
\vfill
\end{document}